\documentclass{iet-ell}

\usepackage{algorithmic}
\usepackage{algorithm}
\usepackage{color}

\setcopyright{open}
\ietVolume{00}
\ietYear{2021}
\ietDoi{ell2.10001}

\received{10 January 2021}
\accepted{4 March 2021}
\usepackage{hyperref}
\begin{document}

\title{Learning from Mixed Datasets: A Monotonic Image Quality Assessment Model}

\author[af1]{Zhaopeng Feng}
\author[af2]{Keyang Zhang}
\author[af3]{Shuyue Jia}
\author[af3]{Baoliang Chen}
\author[af3]{Shiqi Wang}

\affil[af1]{School of Mechanical Engineering and Automation, Harbin Institute of Technology, Shenzhen, China}
\affil[af2]{Department of Electrical Engineering, City University of Hong Kong, Hong Kong SAR}
\affil[af3]{Department of Computer Science, City University of Hong Kong, Hong Kong SAR}

\corresp{Email: shiqwang@cityu.edu.hk}

\begin{abstract}%
Deep learning based image quality assessment (IQA) models usually learn to predict image quality from a single dataset, leading the model to overfit specific scenes. To account for this, mixed datasets training can be an effective way to enhance the generalization  capability of the model. However, it is nontrivial to combine different IQA datasets, as their quality evaluation criteria, score ranges, view conditions, as well as subjects are usually not shared during the image quality annotation. In this paper, instead of aligning the annotations, we propose a monotonic neural network for IQA model learning with different datasets combined. In particular, our model consists of a dataset-shared quality regressor and several dataset-specific quality transformers. The quality regressor aims to obtain the perceptual qualities of each dataset while each quality transformer maps the perceptual qualities to the corresponding dataset annotations with their monotonicity maintained. The experimental results verify the effectiveness of the proposed learning strategy and our code is available at https://github.com/fzp0424/MonotonicIQA.
\end{abstract}

\maketitle%

\section{Introduction}\label{sec:Introduction}
Image quality assessment (IQA) model which aims to provide image quality objectively has been widely used as a key component in various computer vision tasks \cite{liu2017quality}. Popular IQA models can be classified into full-reference IQA (FR-IQA) models and no-reference (NR-IQA) models, depending on whether the reference image is available. FR-IQA models can be dated back to the 1970s when the signal fidelity measures are adopted for quality prediction, such as Peak Signal-to-Noise Ratio (PSNR), Structural Similarity Index (SSIM) \cite{wang2004image}, and {Multi-scale} SSIM (MS-SSIM) \cite{wang2003multiscale}. Compared with FR-IQA, NR-IQA is regarded as a more challenging task due to the lack of reference information. Typical NR-IQA models rely on the natural scene statistics (NSS) construction  \cite{moorthy2011blind, saad2012blind, hou2014blind} with the assumption that the image quality can be estimated by measuring the destruction level of the NSS. In recent years, many pioneers resort to deep learning technology for both FR-IQA \cite{zhang2018unreasonable,bosse2017deep,kim2017deep, ding2020image, chen2022no} and NR-IQA \cite{kang2014convolutional,bosse2017deep,gu2019blind,fu2016blind,kim2018multiple}, endowing the IQA models with a strong capability for quality-aware feature extraction. In particular, a shallow ConvNet was adopted for patch-based NR-IQA learning
in \cite{kang2014convolutional}. Both the quality estimation and distortion identification were learned in a multi-task manner for NR-IQA in \cite{kang2015simultaneous}. The rank information between image pairs was utilized in \cite{liu2017rankiqa,niu2019siamese,chen2021no} which was able to  enrich the training data efficiently. 

\begin{figure}[t]
\centering{\includegraphics[width=0.95\columnwidth]{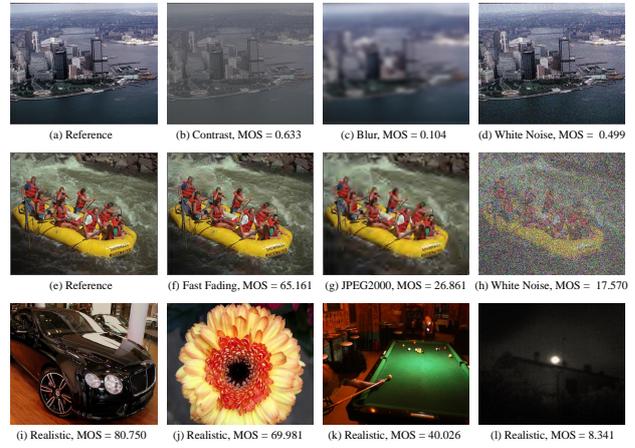}}
\caption{Different IQA datasets consist of images with different contents, distortion types, and score ranges. The distortion type and MOS value are shown below each sub-image. (a)-(d) Images sampled from CSIQ \cite{larson2010most} and (a) is the reference image. (e)-(h) Images sampled from LIVE \cite{sheikh2006statistical} and (e) is the reference image. (i)-(l) Images sampled from KonIQ-10k \cite{hosu2020koniq}}. 
\label{intro}
\end{figure}

Generally speaking, the performance of deep learning based methods heavily relies on the training data. An ideal training set should contain sufficiently diverse image contents, plenty of distortion types, and reliable annotations. However, it is an extremely laborious task to collect and construct such a dataset, leading to the constrained size of existing IQA datasets. The lack of training data usually brings the over-fitting problem and leads to the data-driven methods with poor generalization capability. An efficient way to enrich the training data is to combine the existing datasets together for training. As shown in Fig.~\ref{intro}, the image contents and distortion types in different IQA datasets can be diverse, implying the promising potential for generalized IQA model learning when they are mixed together. Despite the advantages, the quality annotations of existing datasets are usually not fully aligned. As shown in Fig.~\ref{intro}, the mean opinion scores (MOSs) range of the CSIQ dataset \cite{larson2010most} is from 0 to 1 while the range is $[0,100]$ in the LIVE \cite{sheikh2006statistical} dataset. Herein, one can re-scale the MOSs of each dataset non-linearly for training, while the re-scaling function is difficult to be acquired reliably. To account for this,  Zhang \textit{et al.} proposed to use the pairwise rank information in each dataset for mixed datasets training \cite{zhang2021uncertainty}. However, the number of rank pairs will increase explosively with the image number, causing the model convergence for a long-time. In \cite{zhang2022continual}, continual learning was adopted for diverse datasets learning  while the catastrophic forgetting problem may not be completely avoided. Ma \textit{et al.} proposed an incremental learning scheme for cross-task NR-IQA~\cite{ma2021remember}. However, the pruning operator potentially causes performance degradation on the old task. In comparison, we learn the model from all the datasets simultaneously, achieving a better performance balance among different datasets.

Inspired by the four-parameter logistic function,  Li \textit{et al.}  proposed a non-linear mapping layer to map the relative quality to the perceptual quality in \cite{li2021unified}, while the fitting capability of the non-linear function might be questioned.

In this paper, we develop a new paradigm for mixed dataset training by introducing a monotonic neural network. In particular, we first adopt a dataset-shared network for perceptual quality regression of each image and subsequently map the perceptual quality monotonously to the annotated MOSs with the quality ranking in each dataset. The proposed training strategy enjoys two desired advantages.  First, it gets rid of the laborious quality re-alignment to successfully combine different datasets for training. Second, the learnable monotonic neural network possesses a powerful {non-linear} fitting capability to map the image perceptual quality to the annotations. We evaluate our method on six IQA datasets that extensively cover both synthetic distortions and  realistic distortions. Compared with state-of-the-art NR-IQA models, our method presents superior performance for image quality prediction, revealing the effectiveness of the proposed monotonic neural network.
\begin{figure}[t]
\centering{\includegraphics[width=1.0\columnwidth]{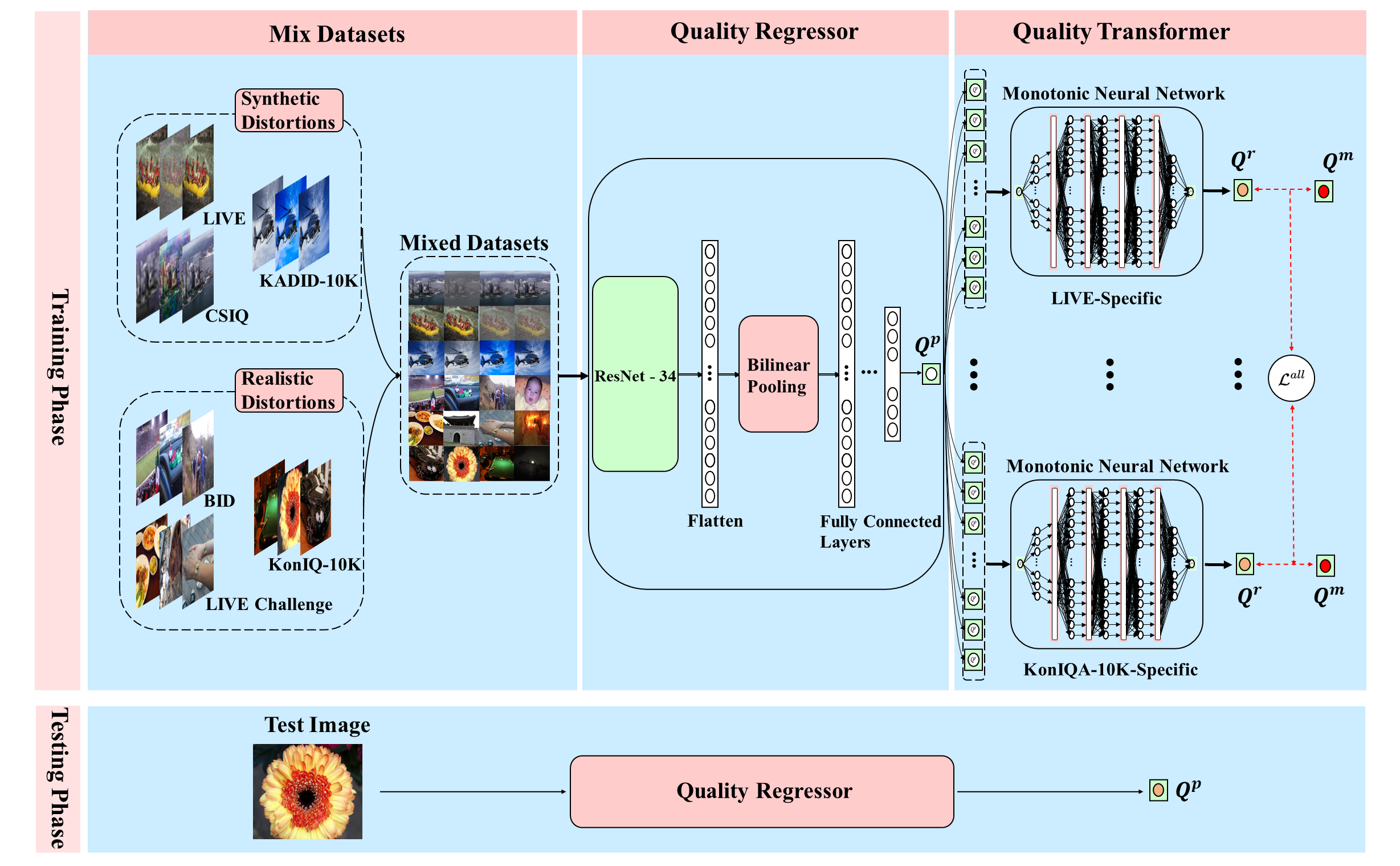}}
\caption{The overview framework of our proposed method. It contains two main components: dataset-shared quality regressor and dataset-specific quality transformer.  In the testing phase, only the quality regressor is used no matter which dataset the testing image is from.}
\label{model}
\end{figure}
\section{Proposed Method}
The overview framework of our method is presented in Fig.~\ref{model}. As shown in Fig.~\ref{model}, our method contains two main components: 1) the dataset-shared quality regressor which aims to obtain the perceptual quality of each image in the mixed datasets. 2) the dataset-specific quality transformer which monotonously maps the perceptual quality to the MOSs annotated in each dataset. The details of each component are elaborated.
\begin{figure}[t]
\centering{\includegraphics[width=0.8\columnwidth]{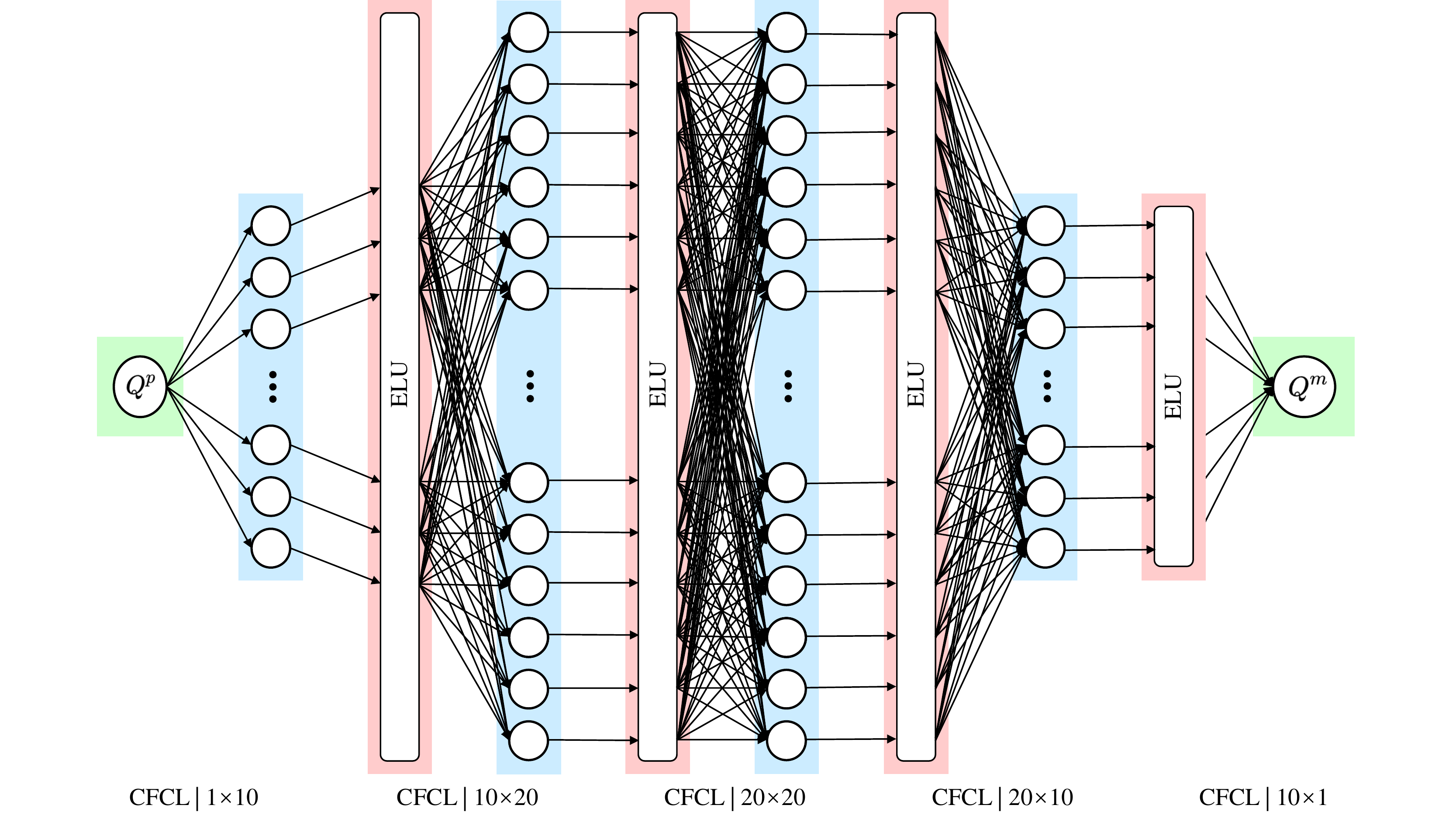}}
\caption{The architecture of our monotonic neural network. We denote the constrained fully connected layer as CFCL and denote the parameterization of CFCL as ``input features $\times$ output features".
\label{transformer}}
\end{figure}

\textit{Quality Regressor.} For the perceptual quality regression, we first adopt the pre-trained ResNet-34 \cite{he2016deep} to extract the quality-aware feature of each image. Then we flatten the spatial dimensions of the feature representation after the last convolution to obtain the feature $\bf{F} \in  \mathbb{R}^{c \times l}$, where $c$ is the channel number and $l$ is the spatial length. Analogous to \cite{zhang2021uncertainty}, we perform the bilinear pooling on $\bf{F}$, aiming to acquire the interactions between different channels,
\begin{align}
     \bf{\overline{F}} = F\times F^T,
\end{align}
where $\bf{\overline{F}} \in  \mathbb{R}^{c \times c}$. Based upon $\bf{\overline{F}}$, two fully connected layers (each followed by a ReLU activation) with a final fully connected layer (the output dimension is one) appended are utilized to regress the $\bf{\overline{F}}$ to the perceptual quality ${Q^p}$. The quality regressor is shared by different datasets. 

\textit{Quality Transformer.} The connection between the perceptual quality ${Q^p}$ of each image and its MOS ${Q^{m}}$ is that their rank should be the same within the dataset. Following this principle, we propose a monotonic neural network to map the ${Q^p}$ to the ${Q^{m}}$. The architecture of our monotonic neural network is shown in Fig.~\ref{transformer}. It consists of five constrained fully connected layers (CFCL) whose weights are all constrained to be positive. The positive weight ensures the gradient of the mapping function to be positive \textit{a.k.a} higher  ${Q^p}$ will be mapped to the higher  ${Q^{m}}$ and vice versa.  In addition,  to maintain the network monotonicity and enhance the transformer non-linear fitting ability, we adopt the ELU \cite{clevert2015fast} as the activation function. We provide a proof of the monotonicity of the quality transformer at \textit{\url{https://drive.google.com/file/d/1Ogwv6y2sSgvOAy_s53gaKYAL_BtNP3Gx/view?usp=share_link}} .

\textit{Loss Function.}  We denote the mapping result (output) of the quality transformer as ${Q^r}$ and the smooth $l_1$ loss $\mathcal{L}^{sl}$ and normal-in-normal loss $\mathcal{L}_{nin}$  \cite{li2020norm} are adopted for training as follows,
\begin{equation}
      \mathcal{L}_{all} =  \mathcal{L}_{sl} (Q^r, Q^{m}) + {\lambda} \mathcal{L}_{nin} (Q^r, Q^{m}),
      \label{all}
\end{equation}
where ${\lambda}$ is a hyper-parameter and
\begin{align}
 \mathcal{L}_{sl} (Q^r, Q^{m}) =
\begin{cases}
  \frac{1}{2}(Q^r-Q^{m})^{2} & \text{ if } \left | Q^r-Q^{m} \right | <1, \\
  \left | Q^r-Q^{m}\right |-\frac{1}{2} & \text{ otherwise} ,
\end{cases}
\end{align}
and
\begin{align}
 \mathcal{L}_{nin} (Q^r, Q^{m}) = \frac{1}{2}\left(\frac{Q^r-\mu ^r}{\sigma ^r}-\frac{Q^{m}-\mu ^{m}}{\sigma^{m}}\right)^{2}, 
\end{align}
with the $\mu ^r$, $\sigma ^r$, $\mu ^{m}$, and $\sigma^{m}$are the mean and standard deviation of the predicted quality scores and MOSs in a training batch.  Our model is trained in an end-to-end manner.

\begin{table*}[h]
\centering
\caption{The median SRCC and PLCC of ten splitting results. The databases used for training models are included in the bracket.} 
\label{table1}  
\resizebox{0.85\textwidth}{!}{%
\begin{tabular}{lccccccclccccccc}
\hline
& \multicolumn{7}{c}{SRCC}                                                                    &  & \multicolumn{7}{c}{PLCC}                                                                 \\ \cline{2-8} \cline{10-16} 
Database                                       & LIVE           & CSIQ           & KADID-10k      & BID            & LIVE Challenge & KonIQ-10k      & Weighted       &  & LIVE           & CSIQ           & KADID-10k      & BID            & LIVE Challenge & KonIQ-10k      & Weighted       \\ \hline
NIQE                                           & 0.906          & 0.632          & 0.374          & 0.468          & 0.464          & 0.521          & 0.471          &  & 0.908          & 0.726          & 0.428          & 0.461          & 0.515          & 0.529          & 0.503          \\
ILNIQE                                         & 0.907          & 0.832          & 0.531          & 0.516          & 0.469          & 0.507          & 0.528          &  & 0.912          & 0.873          & 0.573          & 0.533          & 0.536          & 0.534          & 0.576          \\
PQR (BID)                                       & 0.663          & 0.522          & 0.321          & -              & 0.691          & 0.614          & -              &  & 0.673          & 0.612          & 0.403          & -              & 0.740          & 0.650          & -              \\
PQR (KADID-10k)                                 & 0.916          & 0.803          & -              & 0.358          & 0.439          & 0.485          & -              &  & 0.927          & 0.870          & -              & 0.405          & 0.501          & 0.484          & -              \\
DB-CNN (CSIQ)                                   & 0.855          & -              & 0.501          & 0.329          & 0.451          & 0.499          & -              &  & 0.854          & -              & 0.569          & 0.383          & 0.472          & 0.515          & -              \\
DB-CNN (LIVE Challenge)                         & 0.723          & 0.691          & 0.488          & 0.809          & -              & 0.770          & -              &  & 0.754          & 0.685          & 0.529          & 0.832          & -              & 0.825          & -              \\
KonCept512 (KonIQ-10k)                          & 0.776          & 0.639          & 0.491          & 0.800          & 0.781          & \textbf{0.917} & 0.800          &  & 0.772          & 0.663          & 0.508          & 0.818          & 0.844          & \textbf{0.931} & 0.727          \\
UNIQUE-Linear (All databases) & 0.935          & 0.821          & 0.870          & 0.809          & 0.799          & 0.868          & 0.865          &  & -              & -              & -              & -              & -              & -              & -              \\
UNIQUE (All databases)                          & \textbf{0.969} & \textbf{0.902} & \textbf{0.878} & \textbf{0.858} & \textbf{0.854} & 0.896          & \textbf{0.888} &  & \textbf{0.968} & \textbf{0.927} & \textbf{0.876} & \textbf{0.873} & \textbf{0.890} & 0.901          & \textbf{0.892} \\
Proposed (All databases)                         & \textbf{0.958} & \textbf{0.898} & \textbf{0.885} & \textbf{0.847} & \textbf{0.854} & \textbf{0.902} & \textbf{0.892} &  & \textbf{0.965} & \textbf{0.920} & \textbf{0.887} & \textbf{0.862} & \textbf{0.886} & \textbf{0.911} & \textbf{0.900} \\ \hline
\end{tabular}%
}
\end{table*}

\begin{table}[h]
\centering
\caption{The median SRCC of ten splitting results using different structure of the proposed monotonic neural network.} 
\label{ablation}  
\resizebox{\columnwidth}{!}{%
\begin{tabular}{lccccccc}
\hline
SRCC                                & LIVE  & CSIQ  & KADID-10k & BID   & LIVE Challenge & KonIQ-10k & Weighted \\ \hline
Three CFCLs and Two ELUs            & 0.963 & 0.821 & 0.806     & \textbf{0.862} & 0.851          & \textbf{0.903}     & 0.857    \\
Five CFCLs and Four ELUs (Proposed) & 0.958 & \textbf{0.898} & \textbf{0.885}     & 0.847 & 0.854          & 0.902     & \textbf{0.892}    \\
Seven CFCLs and Six ELUs            & \textbf{0.965} & 0.812 & 0.834     & 0.860 & \textbf{0.855}          & 0.902     & 0.868    \\ \hline
\end{tabular}%
}
\end{table}

\section{Experiments}
To verify the effectiveness of our method, six IQA databases are used and mixed for training, including  LIVE \cite{sheikh2006statistical}, CSIQ 
 \cite{larson2010most}, KADID-10k \cite{lin2019kadid}, BID \cite{ciancio2010no}, LIVE Challenge \cite{ghadiyaram2015massive} and KonIQ-10k \cite{hosu2020koniq}.  The synthetic distortions are involved in the first three  datasets and the last  three  datasets contain the realistic distortions.
We randomly split each database into three subsets for training, validation, and testing. There is no content overlap among the three subsets.  The split proportions are 60$\%$, 20$\%$, and 20$\%$, receptively.  We apply Spearman's rank-order correlation coefficient (SRCC) and Pearson linear correlation coefficient (PLCC) to evaluate the monotonicity and accuracy. The larger SRCC and PLCC indicate better quality prediction results. We linearly re-scale the MOS ranges of each dataset to $[0,10]$. We implement our method by PyTorch \cite{paszke2019pytorch} with Adam \cite{kingma2014adam} as the optimizer. The learning rates of the quality regressor and quality transformer are $3\times10^{-5}$ and $3\times10^{-4}$, respectively. {The learning rates of the pre-trained ResNet-34 and the rest of the model are initialized to $3\times10^{-5}$ and $3\times10^{-4}$, respectively.} In the training phase, we set the batch size by 32 and the ${\lambda}$ in Eqn.~\eqref{all} is 1.0. We first re-scale the short length of images in the training sets and test sets to 512  and  {768} respectively, with their  aspect ratios maintained.  Then patches with a size of  $384 \times 384 \times 3$ are cropped from each training sample for data augmentation during training.


\textit{Performance Comparisons.} We compare our method with six state-of-the-art NR-IQA models, including NIQE \cite{mittal2012making}, ILNIQE \cite{zhang2015feature}, PQR \cite{zeng2018blind}, DB-CNN \cite{zhang2020blind}, KonCept512 \cite{hosu2020koniq} and UNIQUE \cite{zhang2021uncertainty}. The linear re-scaling strategy proposed in \cite{zhang2021uncertainty} is also included and denoted as UNIQUE-Linear \cite{zhang2021uncertainty} in Table~\ref{table1}.
The median SRCC and PLCC  of ten splitting results are reported. For better comparison, the image number weighted SRCC and PLCC of all six datasets are also presented.
As shown in Table~\ref{table1}, we can observe that the NSS-based methods including NIQE \cite{mittal2012making} and ILNIQE \cite{zhang2015feature} achieve  better performance on synthetic datasets while are less effective on realistic datasets. The performance of deep learning based methods is determined by the training set. For example, when the DB-CNN \cite{zhang2020blind} is trained on the CSIQ \cite{larson2010most} dataset, higher SRCC can be achieved on the LIVE \cite{sheikh2006statistical} and KADID-10k \cite{lin2019kadid} datasets than it on the LIVE Challenge \cite{ghadiyaram2015massive}, BID \cite{ciancio2010no}, and KonIQ-10k \cite{hosu2020koniq} datasets. On the contrary, the performance on synthetic datasets will drop dramatically when the  KonIQ-10k \cite{hosu2020koniq} is used for training. Compared with the models that are trained on a single dataset, significant performance improvement can be observed by UNIQUE \cite{zhang2021uncertainty} and our method, revealing the effectiveness of the mixed dataset training.  Moreover, our method  achieves the best performance in terms of both the weighted SRCC and PLCC, demonstrating the superiority of the monotonic neural network based  learning strategy when compared with the linear re-scaling strategy (UNIQUE-Linear) or the rank-based strategy (UNIQUE).

As shown in  Table~\ref{ablation}, we further conduct extra experiments to study the best number of CFCL in our proposed monotonic neural network. From Table~\ref{ablation}, we can observe the best layer number is five. Less or more layers will cause the  performance to drop to some extent. The reason may lie in that fewer layers will bring a negative effect on the non-linear fitting ability and more layers will cause the overfitting problem. To verify our model efficiency, we further compare the floating point operations (FLOPs) that our method consumed during inference with the FLOPs of the other three deep-learning based methods, including the KonCept512, DB-CNN, and UNIQUE. The results show that our method is comparable with the  UNIQUE (3.95G \textit{v.s} 3.68G) and presents a significant gain compared with DB-CNN (16.53G) and KonCept512 (13.22G). It reveals that our method achieves better quality prediction performance without much sacrifice of model complexity.

\section{Conclusion}
In this paper, we concentrate on the NR-IQA model learning from mixed datasets. To avoid the laborious quality annotation alignment,  a monotonic neural network is proposed to map the regressed quality to the MOS values. Experimental results demonstrate that the proposed method is more effective  than both the  unreliable linear re-scaling strategy and  pair rank strategy. We believe our method is able to shed the light on the exploration of more generalized IQA models and inspire more works for learning monotonic neural networks.

\balance

\nocite{*}
\bibliography{iet-ell}
\bibliographystyle{iet}

\end{document}